\documentclass{article}
\usepackage{ijcai24}

\usepackage{times}
\usepackage{soul}
\usepackage{url}
\usepackage[hidelinks]{hyperref}
\usepackage[utf8]{inputenc}
\usepackage[small]{caption}
\usepackage{graphicx}
\usepackage{amsmath}
\usepackage{amsthm}
\usepackage{booktabs}
\usepackage{algorithm}
\usepackage{algorithmic}
\usepackage[switch]{lineno}

\usepackage{xcolor}
\usepackage{listings}
\usepackage{caption}
\usepackage{wasysym}
\usepackage{amsfonts}
\usepackage{tikz}

\definecolor{codegreen}{rgb}{0,0.6,0}
\definecolor{codegray}{rgb}{0.5,0.5,0.5}
\definecolor{codepurple}{rgb}{0.58,0,0.82}
\definecolor{backcolour}{rgb}{0.98,0.98,0.97}
\definecolor{backcolour1}{rgb}{1,1,1}

\makeatletter
\pgfdeclareshape{document}{
\inheritsavedanchors[from=rectangle] %
\inheritanchorborder[from=rectangle]
\inheritanchor[from=rectangle]{center}
\inheritanchor[from=rectangle]{north}
\inheritanchor[from=rectangle]{south}
\inheritanchor[from=rectangle]{west}
\inheritanchor[from=rectangle]{east}
\backgroundpath{%
\southwest \pgf@xa=\pgf@x \pgf@ya=\pgf@y
\northeast \pgf@xb=\pgf@x \pgf@yb=\pgf@y
\pgf@xc=\pgf@xb \advance\pgf@xc by-10pt %
\pgf@yc=\pgf@yb \advance\pgf@yc by-10pt
\pgfpathmoveto{\pgfpoint{\pgf@xa}{\pgf@ya}}
\pgfpathlineto{\pgfpoint{\pgf@xa}{\pgf@yb}}
\pgfpathlineto{\pgfpoint{\pgf@xc}{\pgf@yb}}
\pgfpathlineto{\pgfpoint{\pgf@xb}{\pgf@yc}}
\pgfpathlineto{\pgfpoint{\pgf@xb}{\pgf@ya}}
\pgfpathclose
\pgfpathmoveto{\pgfpoint{\pgf@xc}{\pgf@yb}}
\pgfpathlineto{\pgfpoint{\pgf@xc}{\pgf@yc}}
\pgfpathlineto{\pgfpoint{\pgf@xb}{\pgf@yc}}
\pgfpathlineto{\pgfpoint{\pgf@xc}{\pgf@yc}}
}
}
\makeatother

\DeclareCaptionFormat{mylst}{\hrule#1#2#3}
\newtheorem{example}{Example}

\newcommand{\pishield}{PiShield}

\definecolor{darkgreen}{rgb}{0.0, 0.5, 0.0}
\definecolor{darkred}{rgb}{0.7, 0.0, 0.0}

\title{\pishield: A PyTorch Package for Learning with Requirements}

\author{
Mihaela C\u{a}t\u{a}lina Stoian$^1$
\and
Alex Tatomir$^1$\and
Thomas Lukasiewicz$^{2,1}$\And
Eleonora Giunchiglia$^2$\\
\affiliations
$^1$Department of Computer Science, University of Oxford, UK\\
$^2$Institute of Logic and Computation, Vienna University of Technology, Austria\\
}

\begin{document}

\maketitle

\begin{abstract}
Deep learning models 
have shown their strengths in various application domains, however, they often struggle to meet safety requirements for their outputs.
In this paper, we introduce \pishield{}, the first package ever allowing for the integration of the requirements into the neural networks' topology.  
\pishield{} guarantees compliance with these requirements, regardless of input.
Additionally, it allows for integrating requirements both at inference and/or training time, depending on the practitioners' needs. 
Given the widespread application of deep learning, there is a growing need for frameworks allowing for the integration of the requirements across various domains. Here, we explore three application scenarios: functional genomics, autonomous driving, and tabular data generation.

\end{abstract}

\section{Introduction}

Deep neural networks (DNNs) have shown their strengths in various application domains.
However, they often fail to comply with given requirements defining the safe output space of the model.
To obviate this problem, neuro-symbolic AI methods were introduced, which can be broadly classified into two categories.
The first comprises methods able to integrate the requirements in the loss function and penalize the models when they violated the requirements~\cite{diligenti2012SBR,diligenti2017sbrnn,donadello2017ltnsii,xu2018semantic,fischer2019DL2,nandwani2019primal,badreddine2022logic,li2023shortcuts,ahmed2022a,stoian2023tnorm_road}),
and while these approaches help reduce the requirements' violation incidence, they cannot guarantee their satisfaction. 
Emerging later, the second category consists of methods able to incorporate a given set $\Pi$ of requirements (also called constraints) directly in the topology of the network~\cite{giunchiglia2021jair,hoernle2022multiplexnet,ahmed2022spl} and, thus, to guarantee their satisfaction. 
More recently, an alternative method~\cite{krieken2023anesi}, also capable of guaranteeing compliance with the constraints, proposed using neural networks for performing approximate inference in polynomial time to address the scalability problem of probabilistic neuro-symbolic learning frameworks such as DeepProbLog~\cite{manhaeve2018deepproblog}.
For an in-depth survey of the methods combining deep learning with logical constraints, we refer to~\cite{giunchiglia2022ijcai}, while for a broad survey on neuro-symbolic AI, we refer to~\cite{garcez2019survey}.

In this paper, we propose \pishield{}\footnote{Code: \url{https://github.com/mihaela-stoian/PiShield}}$^,$\footnote{Website: \url{https://sites.google.com/view/pishield}}, a PyTorch-based package allowing for seamlessly integrating domain requirements into neural networks by means of new PyTorch layers that can be built on top of any neural network. These layers, which we call Shield Layers, adhere to the principles outlined in our most recent works~\cite{stoian2024cdgm,giunchiglia2024ccnplus}, advocating for a more requirements-driven machine learning as in \cite{giunchiglia2023manifesto}, and guarantee the satisfaction of the requirements regardless of the input.

The Shield Layers can be applied during inference and/or training, depending on the practitioners' needs. For users that have restricted access to a model but require that the outputs of their models are compliant with a set of rules, \pishield{} offers an easy-to-use interface to meet this need. Alternatively, \pishield{} can guide model training, suiting also practitioners accustomed to modifying their models. 
As expected, its ability to ensure compliance makes it ideal for safety-critical scenarios. 
Furthermore,  studies~\cite{giunchiglia2020neurips,giunchiglia2024ccnplus,giunchiglia2021jair,stoian2024cdgm} have repeatedly shown its efficacy in aiding model learning. We illustrate both of these aspects on three different application domains: functional genomics, autonomous driving, and tabular data generation domains.

\paragraph{Related Work.}
A closely related work is Pylon~\cite{ahmed22a_pylon}, a framework built on PyTorch which allows users to integrate constraints into a loss function. 
Similarly, the LTN package~\cite{badreddine2022logic} provides a TensorFlow implementation of the Logic Tensor Networks (LTNs), while LTNTorch~\cite{LTNtorch} provides its PyTorch implementation.
However, unlike our framework, all the methods described above cannot guarantee the satisfaction of the constraints.

\begin{figure}[t]
    \centering
        \vspace{-0.3cm}
    \includegraphics[width=0.69\linewidth]{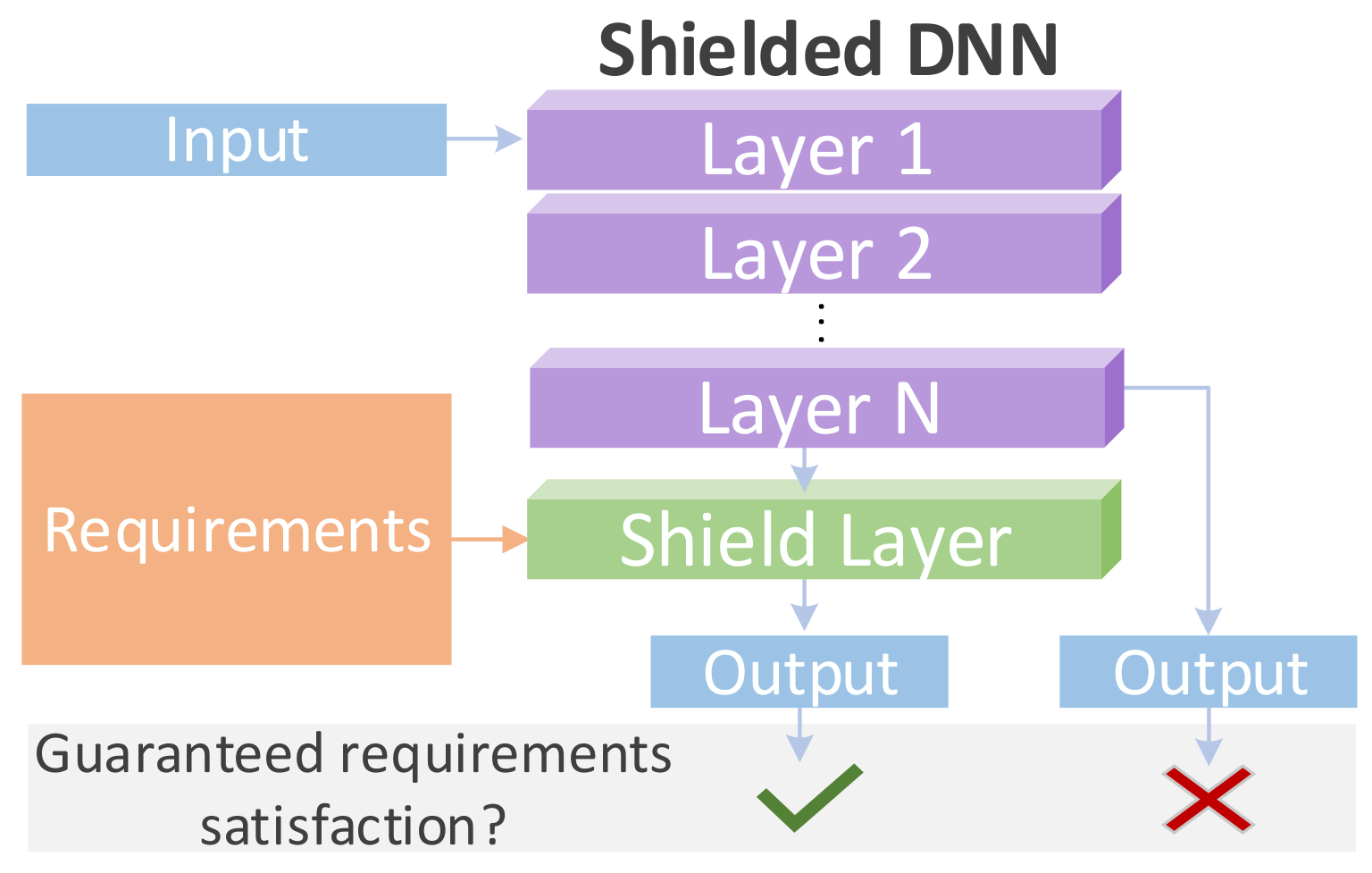}
    \caption{\pishield{} overview.}
    \label{fig:overview}
        \vspace{-0.25cm}
\end{figure}

\section{\pishield{} Overview}

\pishield{} is built on top of PyTorch, allowing for a seamless integration of Shield Layers into neural networks.
Figure~\ref{fig:overview} gives an overview of how to apply \pishield{} to deep neural networks. 
As shown in the Figure, to constrain a DNN during training, the architecture of the DNN needs to be changed by appending a Shield Layer immediately after the output layer.
To use PiShield only at inference time, it is enough to simply build a Shield Layer outside the neural network and apply it on the outputs to make them compliant with the requirements.
Regardless of its use, each Shield Layer requires only two elements to be instantiated: (i) the dimension of the input to the layer (hence in most cases the dimension of the output of the network), and (ii) the path to the file which
contains the requirements.

\paragraph{Requirements.}
The requirements can be expressed either as a propositional logic formula in conjunctive normal form (CNF) or as linear inequalities. In the file, each line should contain a single requirement, which will thus be either a clause 
in the first case or a single linear inequality in the latter.

\begin{example}
Suppose we have a simple multi-label classification problem where we are given as input images taken from an autonomous vehicle, and we have to identify whether a traffic light appears in the image, and whether it is green, yellow, or red. For each image, a standard neural network for this task will output a 4-dimensional vector where each element corresponds to one of the concepts of interest. Suppose, for example, the outputs of the neural networks are ordered in the following way: \texttt{TrafficLight}, \texttt{Red}, \texttt{Yellow}, \texttt{Green}. Then the file, containing the knowledge that a traffic light is always associated with one of the colors and that the colors associated with a traffic light are mutually exclusive, will have the format below:
\begin{center}
    \vspace{-0.05cm}
\begin{tikzpicture}
    \node (example-textwidth-3) [draw, 
                                 shape=document,
                                 text width=0.7\linewidth,    %
                                 align=flush left, 
                                 inner sep=6 pt]%
    {
    \texttt{not y\_0 or y\_1 or y\_2 or y\_3} \\
    \texttt{not y\_0 or not y\_1 or not y\_2} \\
    \texttt{not y\_0 or not y\_1 or not y\_3} \\
    \texttt{not y\_0 or not y\_2 or not y\_3} \\
    };
\end{tikzpicture}
    \vspace{-0.05cm}
\end{center}

\end{example}

\begin{example}
Suppose we have a tabular data generation problem, where we have to generate a synthetic dataset for a clinical trial. Further, suppose that we have the following knowledge about the problem available: (i) the maximum hemoglobin recorded per patient should be always higher or equal than the minimum, and (ii) the maximum temperature recorded should be at least as high as the minimum. 
Then, assuming that the outputs of the neural network are ordered as \texttt{MaxHemoglobin}, \texttt{MinHemoglobin}, \texttt{MaxTemp}, \texttt{MinTemp}, the input file should have the following format:
\begin{center}
\begin{tikzpicture}
    \node (example-textwidth-3) [draw, 
                                 shape=document,
                                 text width=0.5\linewidth,    %
                                 align=flush left, 
                                 inner sep=6 pt]%
    {
    \texttt{y\_0 - y\_1} \texttt{ >= 0}\\
     \texttt{y\_2 - y\_3} \texttt{ >= 0}\\
};
\end{tikzpicture}
\end{center}
    \vspace{-0.25cm}
\end{example}

\begin{figure}[t]
\begin{lstlisting}[label={lis:inference},captionpos=t,language=Python,numbers=left,    numbersep=1pt,caption=Correcting predictions with \pishield{} at inference time.]
from pishield.shield_layer import build_shield_layer

def correct_predictions(predictions, requirements_path):
    num_variables = predictions.shape[-1]
    shield_layer = build_shield_layer(num_variables, requirements_path)
    corrected_predictions = shield_layer(predictions)
    return corrected_predictions
\end{lstlisting}
    \vspace{-0.35cm}
\begin{lstlisting}[label={lis:training},captionpos=t,language=Python,numbers=left,numbersep=1pt, caption=Building a Shield Layer into a DDN to use it for training.]
from pishield.shield_layer import build_shield_layer

class Shielded_DNN(torch.nn.Module):
    def __init__(self, num_dim, requirements_path, ...):
        self.model = torch.nn.Sequential(...)
        self.shield_layer = build_shield_layer(num_dim, requirements_path)
        ...
        
    def forward(self, input):
        output = self.model(input)
        corrected_output = self.shield_layer(output)
        return corrected_output
\end{lstlisting}
\end{figure}

\paragraph{Usage.}
As it can be seen from both Listings~\ref{lis:inference} and~\ref{lis:training}, to build our layer, we only need (lines 5 and 6, respectively) to call the function \texttt{build\_shield\_layer}, which takes as input two parameters: (i) the dimension of the input to the layer, and (ii) the path to the requirements file. Once instantiated, the Shield Layer receives as input a tensor of predictions $p$ (possibly violating the requirements) and returns a tensor $\hat p$ (of the same dimension as $p$), which is now guaranteed to be compliant with the requirements. 
\pishield{} benefits from an easy-to-use interface, and so, correcting $p$ with a Shield Layer is a one-step operation consisting of a forward call of the Shield Layer on $p$.
Listing \ref{lis:inference}, in particular, shows how to do all these steps at inference time. 
On the other hand, Listing~\ref{lis:training} shows how to integrate the layer at training time for which the layer's forward call needs to be done before the backpropagation step.
As shown in the Listing, the easiest way to match this condition is to instantiate the Shield Layer in the constructor of the DNN and then call the layer inside the usual forward method (as shown in line 11).

\section{Example Scenarios}

The need for a package like \pishield{} naturally raises in many application domains. Here, we discuss three example scenarios to which we applied the layers implemented in \pishield{}.

\paragraph{Functional Genomics.}\!\!\!\!\!\!\!\!\!\! In functional genomics, the task is to predict a set of (hierarchical) functions that genes may possess~\cite{vens2008,cerri2018,clare2003}. As such, it belongs to the broad category of Hierarchical Multi-label Classification (HMC) problems, which are multi-label classification problems whose labels are organized in a hierarchical structure.
The hierarchy can be captured by propositional logic rules, where the atoms correspond to the labels. In addition to functional genomics, HMC problems find applications in many real-world domains, such as document (see, e.g.,~\cite{klimt2004,lewis2004}) and image classification (see, e.g.,~\cite{imagenet}), or medical diagnosis (see, e.g.,~\cite{dimitrovski2008}).
In our demo video\footnote{Video: \url{http://tinyurl.com/pdv9eafa}}, we show on a simple task that applying \pishield{} to a model for this task guarantees that the outputs preserve the hierarchical structure.

\begin{figure}[t]
    \centering
    \rotatebox[origin=l]{90}{\centering{\ \ \ \ \ \ \ \ \ Baseline}}
    \includegraphics[width=0.22\textwidth]{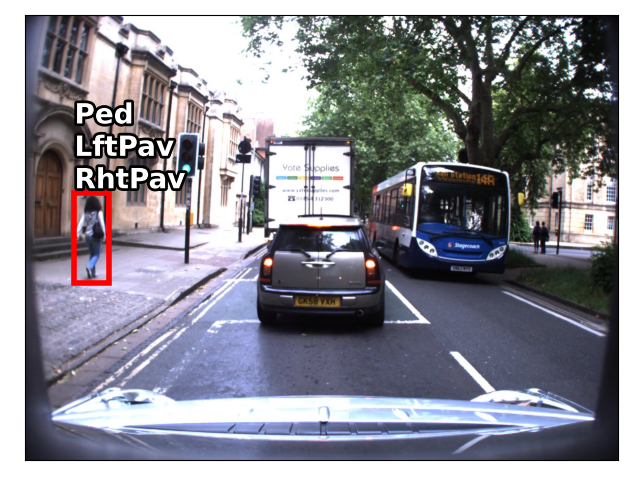}
    \includegraphics[width=0.22\textwidth]{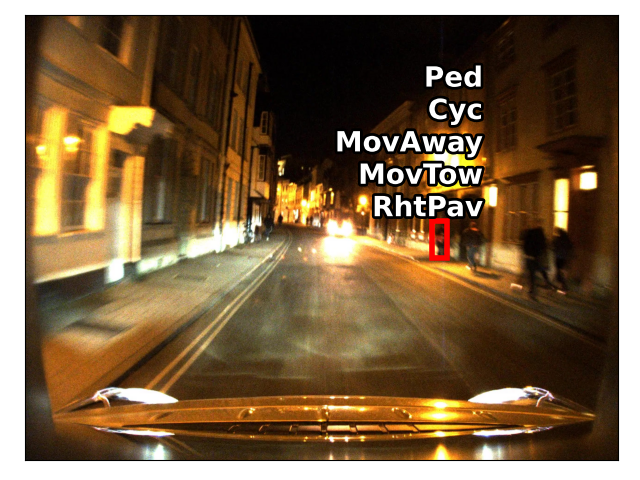}
    \centering
    
    \rotatebox[origin=l]{90}{\ \ \ \ \ \  \ \ \pishield{}}
    \includegraphics[width=0.22\textwidth]{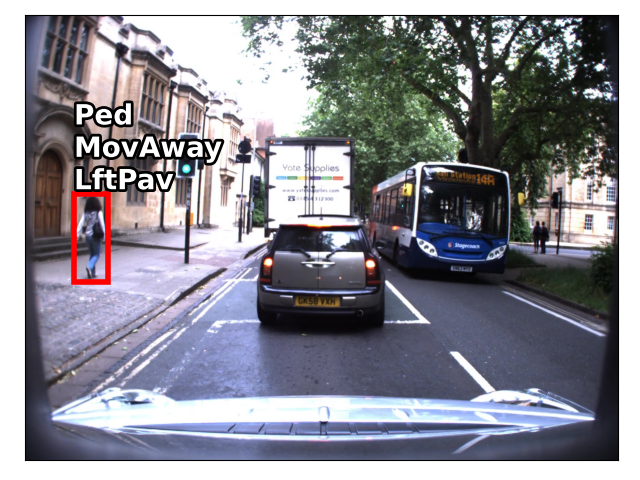}
    \includegraphics[width=0.22\textwidth]{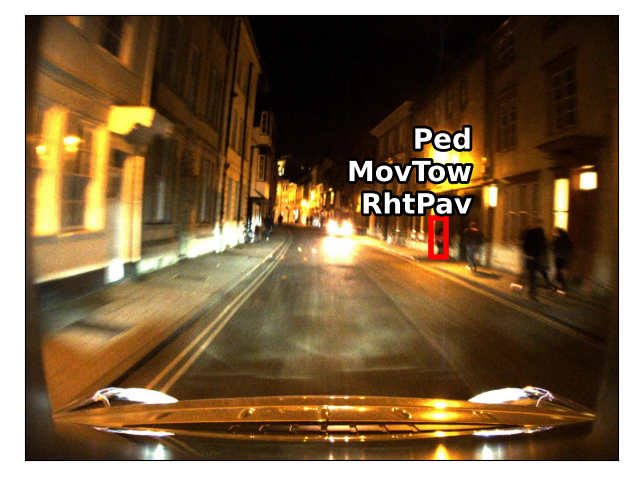}
    \vspace{-0.25cm}
    \caption{The unconstrained models (top row) violate simple background knowledge rules, predicting (i) that a pedestrian is both on the right and on the left pavement, and (ii) that a person is both a pedestrian and a cyclist, and is moving towards and away from the self-driving vehicle at the same time. On the other hand, the predictions made using \pishield{} (bottom row) are guaranteed to be compliant with the background knowledge. }
    \label{fig:roadr_left_right}
\end{figure}

\paragraph{Road Events Detection.} 
Generalizing the above problem, we have the task of assigning  sets of labels to instances of various types, including images, documents etc.
Given available domain knowledge, constraints are placed on the possible outputs, eliminating impossible (i.e., contradicting background knowledge) outcomes.
In our recent work~\cite{giunchiglia2024ccnplus}, we considered the task of multi-label classification for an autonomous driving scenario using the ROAD-R dataset~\cite{giunchiglia2022road}, the first real-world dataset for autonomous driving with manually annotated logical constraints.
ROAD-R is build upon the road event detection dataset ROAD~\cite{road} by adding 243 propositional logic requirements written in CNF.
We showed that applying Shield Layers on the outputs is not only able to guarantee that these requirements are satisfied, but also results in increased performance, as seen in Table 2 of \cite{giunchiglia2024ccnplus}.
Figure~\ref{fig:roadr_left_right} shows examples of how using \pishield{} can impact the predictions' quality, which is intrinsically coupled with the safety guarantees. %

\begin{figure}[tb]
\centering
    \includegraphics[width=1.\linewidth]{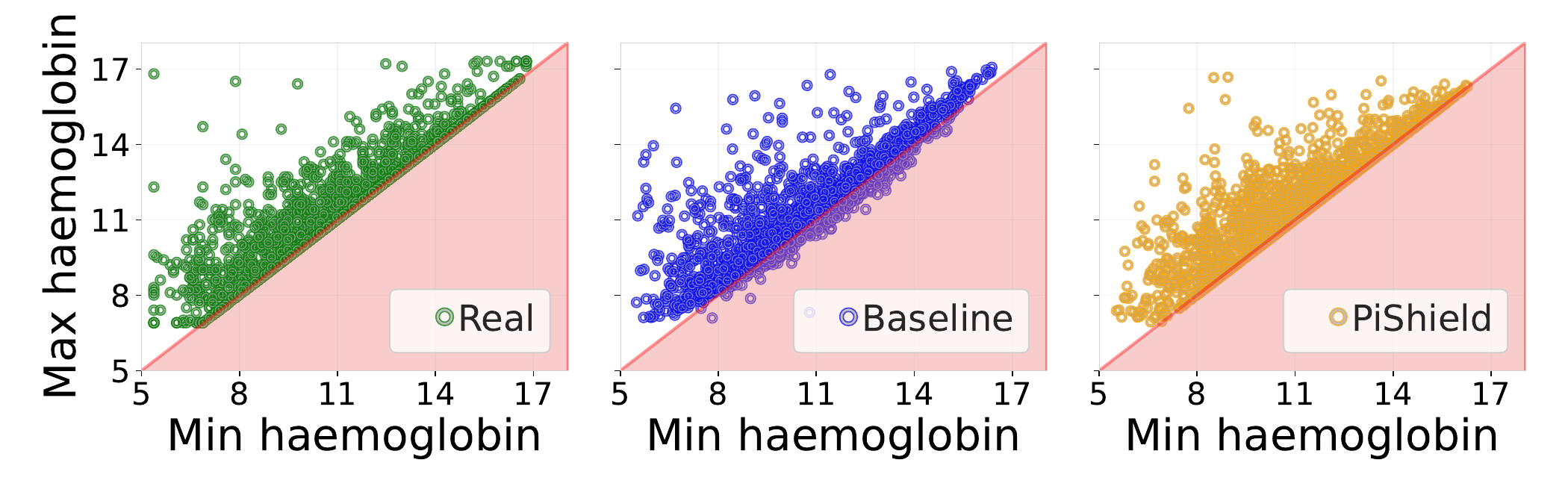}
    \vspace{-0.7cm}
    \caption{Real data (left) and samples generated by an unconstrained neural network (middle) and a neural network constrained with \pishield{} (right).}
    \label{fig:boundaries}
\end{figure}

\begin{table}
    \centering
    \begin{tabular}{lrr}
        \toprule
        Scenario  & Baseline & PiShield \\
        \midrule
        Functional genomics(AU($\overline{\text{PRC}}$)) & 0.225   & \textbf{0.241} \\
        Autonomous driving (f-mAP) & 0.288	& \textbf{0.303} \\
        Tabular data generation (Utility-F1) & 0.430 & \textbf{0.458}\\
        \bottomrule
    \end{tabular}
    \caption{Aggregated performance. The best results are in \textbf{bold}. \hfill \ }
    \label{tab:performance}
\end{table}

\paragraph{Tabular Data Generation.} 
Another application where we utilized \pishield{} is the task of synthesizing tabular data such that the outputs of a model for this task satisfies background knowledge rules provided as linear inequalities, which capture relations between the features of the tabular data. 
Applying Shield Layers to standard deep generative models (DGMs) resulted in more realistic outputs, which are compliant with the background knowledge.
For example, consider the constraint \textit{MaxHaemoglobin $\ge$ MinHaemoglobin}, which is a real constraint that we encountered in \cite{stoian2024cdgm} in one of the datasets. 
In Figure \ref{fig:boundaries}, following the qualitative illustrations in Figures 3-6 of our previous work \cite{stoian2024cdgm}, we show how the constrained DGMs match the real data more closely in the distribution of their outputs.
Additionally, through an extensive empirical analysis, we showed (in Table 2 of \cite{stoian2024cdgm}) that integrating Shield Layers into DGMs yields an increased performance across two standard metrics used in tabular data generation.

\paragraph{Performance.}
In Table~\ref{tab:performance}, we compare baseline models with their constrained versions, which use \pishield{} during training. 
We report results for three real-world applications: (i) functional genomics, based on Table 3 of~\protect\cite{giunchiglia2020neurips}, reporting the area under the average precision and recall curve (AU($\overline{\text{PRC}}$)) over 8 datasets, %
(ii) autonomous driving, based on Table 2 of~\protect\cite{giunchiglia2024ccnplus}, reporting the frame-wise mean average precision (f-mAP) on the ROAD-R dataset, annotated with 243 propositional requirements; 
(iii) tabular data generation, based on Table 2 of~\protect\cite{stoian2024cdgm}, reporting the F1-score for the utility performance (Utility-F1) averaged over 5 deep generative model types and 5 different datasets, annotated with up to 31 linear inequality constraints.
As we can see, using \pishield{} during training provides major performance improvements over the unconstrained baselines. %

\section{Conclusions}

In this paper, we introduced \pishield{}, the first package that allows for injecting requirements into neural networks' topology by building Shield Layers, which correct the outputs so that they are guaranteed to satisfy the rules.
\pishield{} can be applied (i) during training by building Shield Layers into the neural networks' architecture, or (ii) at inference time by correcting the neural networks' outputs as a post-processing step.
We envision our package will be of use to practitioners working on real-world applications where domain knowledge can be expressed as propositional or linear constraints.

\appendix

\section*{Ethical Statement}

There are no ethical issues directly associating with the framework's specification and available methods.
Any ethical issues would only arise from the context in which the framework is planned to be used and on the constraints provided by the users.
For example, when generating tabular data, the users could create a set of constraints that would facilitate (partially) recovering specific records in the training data, thus posing problems in cases where the data generated in this way can be made publicly available, but the training data cannot.

\section*{Acknowledgments}
Mihaela C\u{a}t\u{a}lina Stoian is supported by the EPSRC under the grant EP/T517811/1. 
This work was also supported by the Alan Turing Institute under the EPSRC grant EP/N510129/1, by the AXA Research Fund, by the EPSRC grant EP/R013667/1, and by the EU TAILOR grant.
We would like to thank Salijona Dyrmishi, Thibault Simonetto, Andrew Ryzhikov, Salah Ghamizi, and Maxime Cordy for the useful discussions. 
We also acknowledge the use of the EPSRC-funded Tier 2 facility JADE (EP/P020275/1), GPU computing support by Scan Computers International Ltd.

\bibliographystyle{named}
\bibliography{ijcai24}

\end{document}